\documentclass{article}

\usepackage[nonatbib]{nips_2016}

\usepackage{nips_2016}

\usepackage{biblatex}
\usepackage[utf8]{inputenc} 
\usepackage[T1]{fontenc}    
\usepackage{hyperref}       
\usepackage{url}            
\usepackage{booktabs}       
\usepackage{amsfonts}       
\usepackage{nicefrac}       
\usepackage{microtype}      
\usepackage{algorithm}
\usepackage{algpseudocode}
\usepackage{graphicx}
\usepackage{subcaption}
\usepackage{multicol}

\title{Inverting The Generator Of A Generative Adversarial Network}  

\author{
  Antonia Creswell \\ BICV Group \\ Bioengineering \\ Imperial College London \\
  \texttt{ac2211@ic.ac.uk}\\
  \And
  Anil Anthony Bharath \\ BICV Group \\ Bioengineering \\ Imperial College London \\
}

\begin{document}

\maketitle

\begin{abstract}


Generative adversarial networks (GANs) learn to synthesise new samples from a high-dimensional distribution by passing samples drawn from a latent space through a generative network. When the high-dimensional distribution describes images of a particular data set, the network should learn to generate visually similar image samples for latent variables that are close to each other in the latent space. For tasks such as image retrieval and image classification, it may be useful to exploit the arrangement of the latent space by projecting images into it, and using this as a representation for discriminative tasks. GANs often consist of multiple layers of non-linear computations, making them very difficult to invert. This paper introduces techniques for projecting image samples into the latent space using any pre-trained GAN, provided that the computational graph is available. We evaluate these techniques on both MNIST digits and Omniglot handwritten characters. In the case of MNIST digits, we show that projections into the latent space maintain information about the style and the identity of the digit. In the case of Omniglot characters, we show that even characters from alphabets that have not been seen during training may be projected well into the latent space; this suggests that this approach may have applications in one-shot learning. 



    
\end{abstract}

\section{Introduction}

Generative adversarial networks (GANs) \cite{radford2015unsupervised, goodfellow2014generative} are a class of generative model which are able to generate realistic looking images of faces, digits and street numbers \cite{radford2015unsupervised}. GANs involve training two networks: a generator, $G$, and a discriminator, $D$. The generator, $G$, is trained to generate images from a random vector $z$ drawn from a prior distribution, $P(Z)$. The prior is often chosen to be a normal or uniform distribution.

Radford et al. \cite{radford2015unsupervised} demonstrated that generative adversarial networks (GANs) learn a ``rich linear structure" meaning that algebraic operations in $Z$-space often lead to meaningful generations in image space. Since images represented in $Z$-space are often meaningful, direct access to a $z \in Z$ for a given image, $x \in X$ may be useful for discriminative tasks such as retrieval or classification. Recently, it has also become desirable to be able to access $Z$-space in order to manipulate original images \cite{zhu2016generative}. Further, inverting the generator may provide interesting insights to highlight what the GAN model learns. Thus, there are many reasons that we may want to invert the generator.


Mapping from image space, $X$, to $Z$-space is non-trivial, as it requires inversion of the generator, which is often a many layered, non-linear model \cite{radford2015unsupervised, goodfellow2014generative, chen2016infogan}. Dumoulin et al. \cite{dumoulin2016adversarially} and Donahue et al. \cite{donahue2016adversarial} proposed learning a third, decoder network alongside the generator and discriminator to map image samples back to $Z$-space. Collectively, they demonstrated results on MNIST, ImageNet, CIFAR-10 and SVHN and CelebA. However, reconstructions of inversions are often poor. Specifically, reconstructions of inverted MNIST digits using methods of Donahue et al. \cite{donahue2015long}, often fail to preserve style and character class. Drawbacks to this approach include the need to train a third network which increases the number of parameters that have to be learned, increasing the chances of over fitting. The need to train an extra network also means that inversion cannot be performed on pre-trained networks.

We propose an alternative approach to generator inversion which makes the following improvements:
\begin{itemize}
    \item We infer $Z$-space representations for images that when passed through the generator produce samples that are visually similar to those from which they were inferred. For the case of MNIST digits, our proposed inversion technique ensures that digits generated from inferred $Z$'s maintain both the style and character class of the image from which the $Z$ was inferred, better than those in previous work \cite{donahue2016adversarial}.
    \item Our approach can be applied to a pre-trained generator provided that the computational graph for the network is available.
\end{itemize}  
We also show that batches of $z$ samples can be inferred from batches of image samples, which improves the efficiency of the inversion process by allowing multiple images to be inverted in parallel. In the case where a network is trained using batch normalisation, it may also be necessary to invert a batch of $z$ samples.


Inversion is achieved by finding a vector $z \in Z$ which when passed through the generator produces an image that is very similar to the target image.


\section{Method: Inverting The Generator}

For an image $x \in \Re^{m \times m}$ we want to infer the $Z$-space representation, $z \in Z$, which when passed through the trained generator produces an image very similar to $x$. We refer to the process of inferring $z$ from $x$ as \textit{inversion}. This can be formulated as a minimisation problem:

\begin{equation} \label{cost}
    z^* = \min_z   {- \mathbb E}_x \log[G(z)] 
\end{equation}

Provided that the computational graph for $G(z)$ is known, $z^*$ can be calculated via gradient descent methods, taking the gradient of $G$ w.r.t. $z$. This is detailed in Algorithm \ref{min}.


\begin{algorithm}
\caption{Algorithm for inferring $z^* \in \Re^d$, the latent representation for an image $x \in \Re^{m \times m}$.}\label{min}
\begin{algorithmic}[1]
\Procedure{Infer}{$x$}\Comment{Infer $z^* \in \Re^d$ from $x \in \Re^{m \times m}$}
   \State $z^* \sim P_z(Z)$ \Comment{Initialise z by sampling the prior distribution}
   \While{NOT converged}
      \State $L \gets -( x \log [G(z^*)] + (1-x) \log [1-G(z^*)])$ \Comment{Calculate the error}
      \State $z^*\gets z^* - \alpha \nabla_z L $ \Comment{Apply gradient descent}
   \EndWhile\label{euclidendwhile}
   \State \textbf{return} $z^*$
\EndProcedure
\end{algorithmic}
\end{algorithm}

Provided that the generator is deterministic, each $z$ value maps to a single image, $x$. A single $z$ value cannot map to multiple images. However, it is possible that a single $x$ value may map to several $z$ representations, particularly if the generator collapses \cite{salimans2016improved}. This suggests that there may be multiple possible $z$ values to describe a single image. This is very different to a discriminative model, where multiple images, may often be described by the same representation vector \cite{mahendran2015understanding}, particularly when a discriminative model learns representations tolerant to variations.

The approach of Alg. \ref{min} is similar in spirit to that of Mahendran et al. \cite{mahendran2015understanding}, however instead of inverting a representation to obtain the image that was responsible for it, we invert an image to discover the latent representation that generated it. 

\subsection{Effects Of Batch Normalisation}

GAN training is non-trivial because the optimal solution is a saddle point rather than a minimum \cite{salimans2016improved}. It is suggested by Radford et al. \cite{radford2015unsupervised} that to achieve more stable GAN training it is necessary to use batch normalisation \cite{ioffe2015batch}. Batch normalisation involves calculating a mean and standard deviation over a batch of outputs from a convolutional layer and adjusting the mean and standard deviation using learned weights. If a single $z$ value is passed through a batch normalisation layer, the output of the layer may be meaningless. To prevent this problem, it would be ideal to use virtual batch normalisation \cite{salimans2016improved}, where statistics are calculated over a separate batch. However, we want to allow this technique to be applied to any pre-trained network - where virtual batch normalisation may not have been employed. To counteract the effects of batch normalisation, we propose inverting a mixed batch of image samples at a time. This not only has the desired effect of dealing with problems caused when using batch normalisation, but also allows multiple image samples to be inverted in parallel.


\subsection{Inverting A Batch Of Samples}

Not only does inverting a batch of samples make sense when networks use batch normalisation, it is also a practical way to invert many images at once. We will now show that this approach is a legitimate way to update many $z^*$ values in one go.

Let \textbf{z}$_b \in \Re^{B \times n}$, \textbf{z}$_b=\{z_1, z_2, ... z_B\}$ be a batch of $B$ samples of $z$. This will map to a batch of image samples \textbf{x}$_b \in \Re^{B \times m \times m}$, \textbf{x}$_b=\{x_1, x_2, ... x_B\}$. For each pair $(z_i, x_i)$, $i \in \{1...B\}$, a loss $L_i$, may be calculated. The update for $z_i$ would then be $z_i \gets z_i - \alpha \frac{d L_i}{d z_i}$ 

If reconstruction loss is calculated over a batch, then the batch reconstruction loss would be $\sum_{i=\{1,2...B\}} L_i$, and the update would be: 
\begin{equation}
    \nabla_{\textbf{z}_b} L = \frac{\partial \sum_{i \in \{1,2,...B\}} L_i}{\partial (\textbf{z}_b)}
\end{equation}
\begin{equation}
    =\frac{\partial(L_1+L_2...+L_i)}{\partial(\textbf{z}_b)}
\end{equation}
\begin{equation}
    = \frac{dL_1}{dz_1}, \frac{dL_2}{dz_2}, ... \frac{dL_B}{dz_B}
\end{equation}

Each reconstruction loss depends only on $G(z_i)$, so $L_i$ depends only on $z_i$, which means $\frac{\partial L_i}{\partial z_j}=0$, for all $i\not=j$. Note that this may not strictly be true when batch normalisation is applied to outputs of convolutional layers in the generative model, since batch statistics are used to normalise these outputs. However, provided that the size of the batch is sufficiently large we assume that the statistics of a batch are approximately constant parameters for the dataset, rather than being dependant on the specific $z_{j=1,...B}$ values in the batch. This shows that $z_i$ is updated only by reconstruction loss $L_i$, and the other losses do not contribute to the update of $z_i$, making batch updates a valid approach.

\subsection{Using Prior Knowledge Of P(Z)}

A GAN is trained to generate samples from a $z \in Z$ where the distribution over $Z$ is a chosen prior distribution, $P(Z)$. $P(Z)$ is often a Gaussian or uniform distribution. If $P(Z)$ is a uniform distribution, $\mathcal{U}[a,b]$, then after updating $z^*$, it can be clipped to be between $[a,b]$. This ensures that $z^*$ lies in the probable regions of $Z$. If $P(Z)$ is a Gaussian Distribution, $\mathcal{N}[\mu,\sigma]$, regularisation terms may be added to the cost function, penalising samples that have statistics that are not consistent with $P(Z)=\mathcal{N}[\mu,\sigma]$.

$z \in Z$ is a vector of length $d$. If each of the $d$ values in $z \in \Re^d$ are drawn independently and from identical distributions, and provided that $d$ is sufficiently large, we may be able to use statistics of values in $z$ to add regularisation terms to the loss function. For instance, if $P(Z)$ is a distribution with mean, $\mu$ and standard deviation $\sigma$, we get the new loss function:


\begin{equation} \label{loss}
     L(z,x) = {\mathbb E}_x log[G(z)] + \gamma_1|| \mu - \hat{\mu} ||_2^2 + \gamma_2|| \sigma - \hat{\sigma}||_2^2
\end{equation}

where $\hat{\mu}$ is the mean value of elements in $z$, $\hat{\sigma}$ is the standard deviation of elements in $z$ and $\gamma_{1,2}$ are weights.

Since $d$ is often quite small (e.g. $100$ \cite{radford2015unsupervised}), it is unrealistic to expect the statistics of a single $z$ to match those of the prescribed prior. However, since we are able to update a batch of samples at a time, we can calculate $\hat{\mu}$ and $\hat{\sigma}$ over many samples in a batch to get more meaningful statistics.

\section{Relation to Previous Work}



This approach of inferring $z$ from $x$ bears similarities to work of Zhu et al. \cite{zhu2016generative}; we now highlight the differences between the two approaches and the benefits of our approach over that of Zhu et al. \cite{zhu2016generative}. Primarily, we address issues related to batch normalisation by showing that a mixed batch of image samples can be inverted to obtain latent $z$ encodings. Potential problems encountered when using batch normalisation are not discussed by Zhu et al. \cite{zhu2016generative}.


The generator of a GAN is trained to generate image samples $x \in X$ from a $z \in Z$ drawn from a prior distribution $P(Z)$. This means that some $z$ values are more probable that other $z$ values. It makes sense, then, that the inferred $z$'s are also from (or at least near) $P(Z)$. We introduce hard and soft constraints to be used during the optimisation process, to encourage inferred $z$'s to be likely under the prior distribution $P(Z)$. Two common priors often used when training GAN's are the uniform and normal distribution; we show that our method copes with both of these priors.


Specifically, Zhu et al. \cite{zhu2016generative} calculate reconstruction loss, by comparing the features of $x$ and $G(z^*)$ extracted from layers of AlexNet, a CNN trained on natural scenes. This approach is likely to fail if generated samples are not of natural scenes (e.g. MNIST digits). Our approach considers pixel-wise loss, providing an approach that is generic to the dataset. Further, if our intention is to use the inversion to better understand the GAN model, it may be essential not to incorporate information from other pre-trained networks in the inversion process.




\section{``Pre-trained'' Models}
We train four models on two different datasets, MNIST and Omniglot \cite{lake2015human}. In order to compare the effects of regularisation or clipping when using a normal or uniform prior distribution respectively, we train networks on each dataset, using each prior - totalling four models. 

\subsection{MNIST}
The MNIST dataset consists of $60$k samples of hand written digits, $0$ to $9$. The dataset is split into $50$k samples for training and $10$k samples for testing. Both the training and testing dataset contains examples of digits $0$ to $9$.

The generator and discriminator networks for learning MNIST digits are detailed in Table \ref{MNIST_arch}. The networks were trained for $500$ iterations with batch size $128$, learning rate $0.002$ using Adam updates. The networks are trained on $50$k MNIST training samples, covering all $10$ categories. 

\begin{table}
\centering
\begin{tabular}{cc}\toprule
    G(z) & D(x) \\ \hline
    input: $z \in \Re^{100}$ & input: $x \in \Re^{28 \times 28}$ \\
    fully connected 1024 units + batch norm + relu & conv  64,5,5 + upsample + batch norm + leaky relu(0.2) \\
    fully connected 6272 units + batch norm + relu & conv  128,5,5 + upsample + batch norm + leaky relu(0.2) \\
    reshape(128,7,7) & reshape(6272) \\
    conv 64,5,5 + down-sample + batch norm + relu & fully connected 1024 units + leaky relu(0.2) \\
    conv 1,5,5 + down-sample + batch norm + relu & fully connected 1 unit + leaky relu(0.2)\\
    \bottomrule
\end{tabular}
\caption{\textbf{MNIST Architecture}: Model architecture for learning to generate MNIST characters}
\label{MNIST_arch}
\end{table}

\begin{figure}
\begin{multicols}{2}
    \centering
    \begin{subfigure}{0.5\textwidth}
    \includegraphics[width=0.9\linewidth]{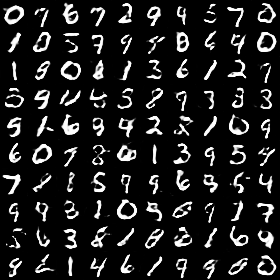}
    \centering
    \caption{MNIST generations with uniform prior.}
    \label{fig:my_label}
    \end{subfigure}
    \begin{subfigure}{0.5\textwidth}
    \includegraphics[width=0.9\linewidth]{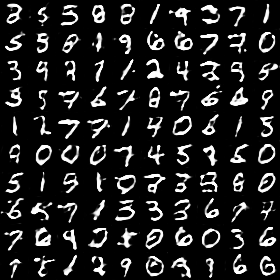}
    \centering
    \caption{MNIST generations with normal prior.}
    \label{fig:my_label}
    \end{subfigure}
    \begin{subfigure}{0.5\textwidth}
    \includegraphics[width=0.9\linewidth]{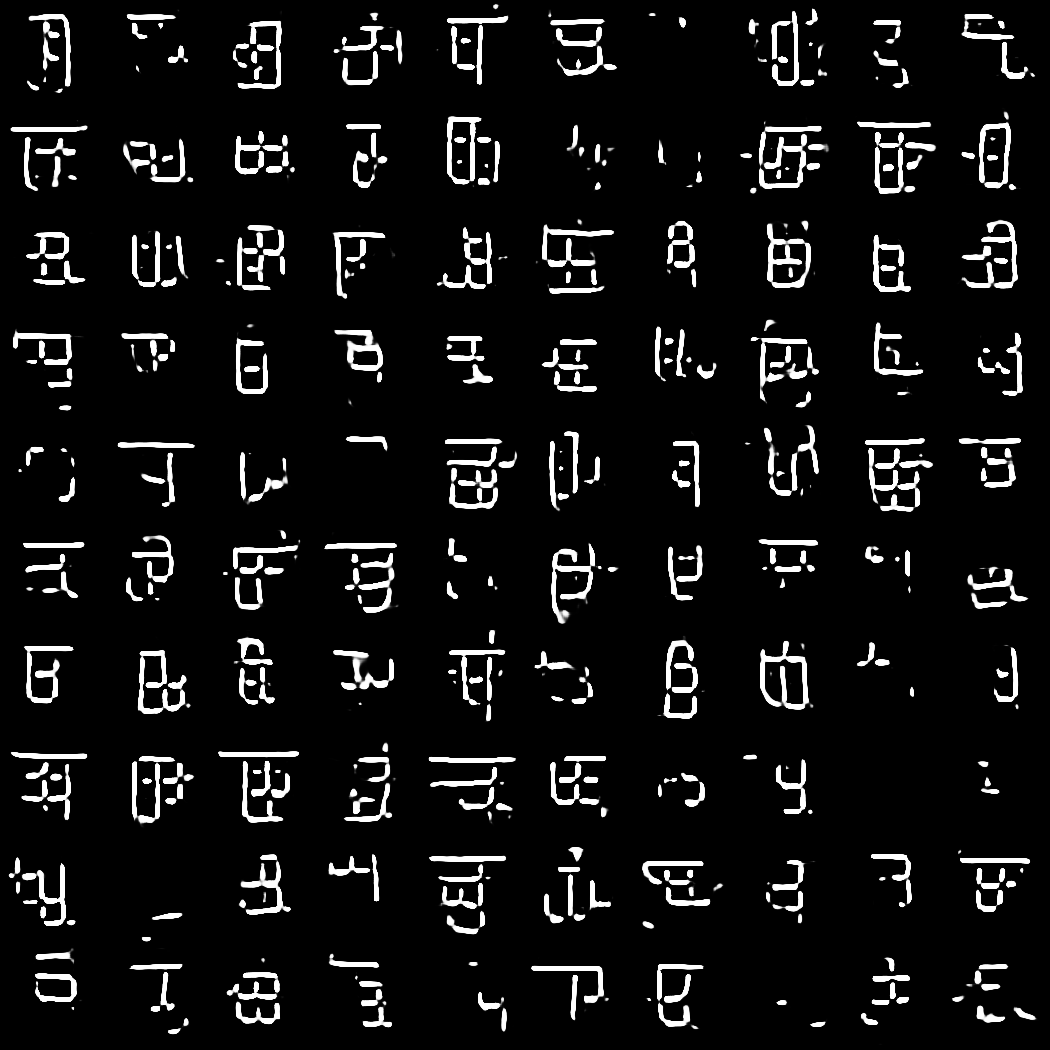}
    \centering
    \caption{Omniglot generations with uniform prior}
    \label{fig:my_label}
    \end{subfigure}
    \begin{subfigure}{0.5\textwidth}
    \includegraphics[width=0.9\linewidth]{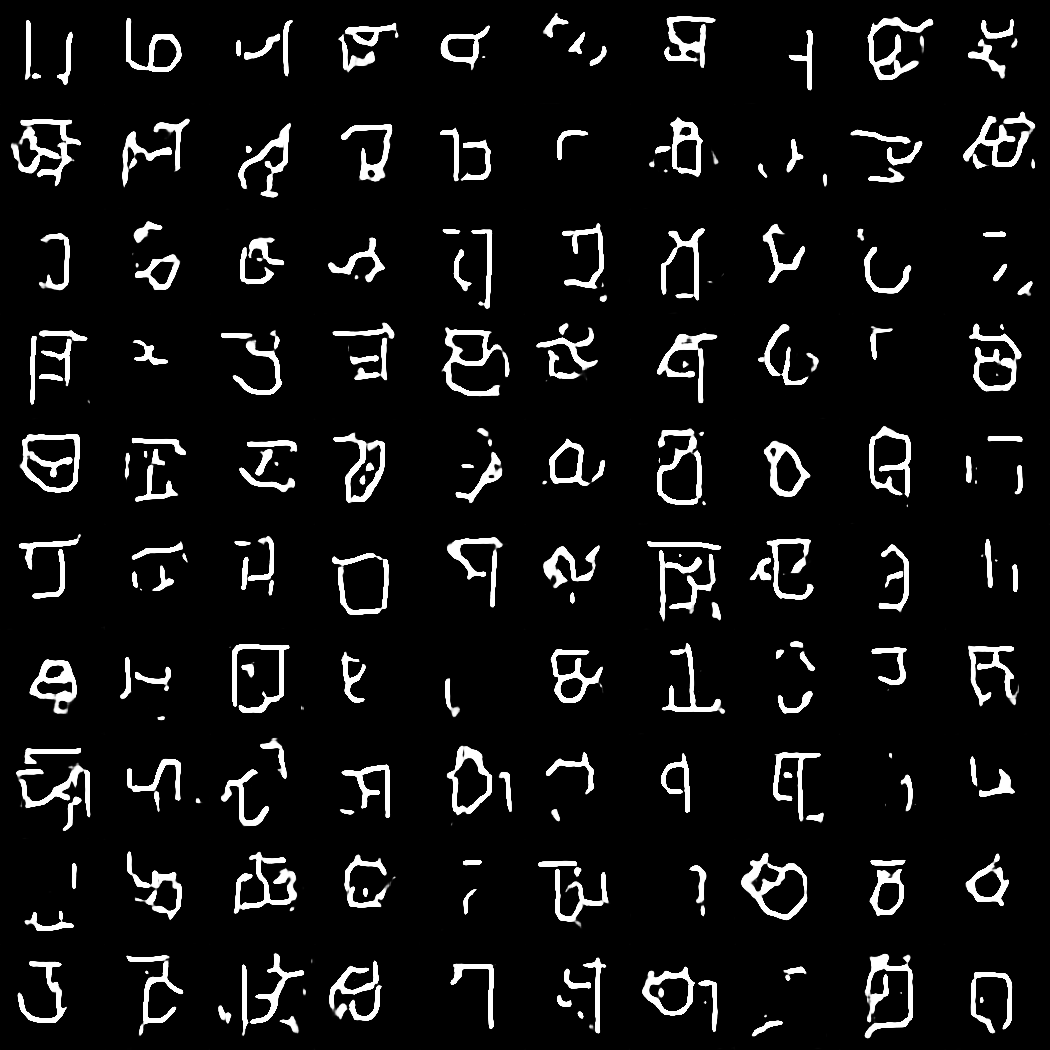}
    \centering
    \caption{Omniglot generations with normal prior}
    \end{subfigure}
\end{multicols}
\centering
\caption{\textbf{Generations for MNIST and Omniglot} using both a Uniform and Normal prior.} 
\label{gen}
\end{figure}

Fig. \ref{gen} shows examples of $100$ random generations for MNIST networks trained using uniform and normal distributions.

\subsection{Omniglot}
The Omniglot dataset \cite{lake2015human} consists of characters from $50$ different alphabets, where each alphabet has at least $14$ different characters. The Omniglot dataset has a background dataset, used for training and a test dataset. The background set consists of characters from $30$ writing systems, while the test dataset consists of characters from the other $20$. Note, characters in the training and testing dataset come from different writing systems. The generator and discriminator networks for learning Omniglot characters \cite{lake2015human} are the same as those used in previous work \cite{creswell2016task}. The network is trained only on the background dataset, for $2000$ iterations with random batches of size $128$, using Adam updates with learning rate $0.002$. The latent encoding has dimension, $d=100$. 

\section{Experiments}
These experiments are designed to evaluate the proposed inversion process. A valid inversion process should map an image sample, $x\in X$ to a $z^* \in Z$, such that when $z^*$ is passed through the generative part of the GAN, it produces an image, $G(z^*)$, that is close to the original image, $x$.

In our experiments, we selected a random batch of images, $x \in X$, and applied inversion to the generator network using this batch. We performed inversion on four generators: Two trained to generate MNIST digits and two trained to generate Omniglot digits.  For each case, the networks were trained to generate from $z \in Z$ with $P(Z)$ being either a uniform or a normal distribution.

To invert a batch of image samples, we minimised the cost function described by Eqn. \ref{cost}. In these experiments we examined the necessity of regularisation or clipping in the minimisation process. If samples may be inverted without the need for regularisation or clipping, then this technique may be considered general to the latent prior, $P(Z)$, used to train the GAN.



\textbf{Minimising Binary Cross Entropy:} We performed inversion where the cost function consisted only of minimising the binary cross entropy between the image sample and the reconstruction. For this approach to be general to the noise process used for latent space, we would hope that image samples may be inverted well by only minimising binary cross entropy and not using any hard or soft constraints on the inferred $z$'s.

\textbf{Regularisation and Clipping:} GANs are trained to generate images from a prior distribution, $P(Z)$. Therefore it may make sense to place some constraints on $z$'s inferred in the inversion process. However, the constraints needed depend on the distribution of the noise source. These experiments deal with two distributions, commonly used when training GANs, the uniform and Gaussian distributions. For generators trained using a uniform distribution we compare inversion with and with out clipping. For generators trained using a Gaussian distribution we compare inversion with and with out regularisation as described by Eqn. \ref{loss}, using $\gamma_{1,2}=1$.
 
\subsection{Evaluation Methods}
To quantitatively evaluated the quality of image reconstruction by taking the mean absolute pixel error across all reconstructions for each of the reconstruction methods. For qualitative evaluation, we show pairs of $x$ and their reconstruction, $G(z^*)$. By visualising the inversions, we can assess to what extent the the digit or character identity is preserved. Also, with the MNIST dataset, we can visually assess whether digit style is also preserved.

\section{Results}

\subsection{MNIST}
\begin{figure}
    \centering
    \includegraphics[width=\textwidth]{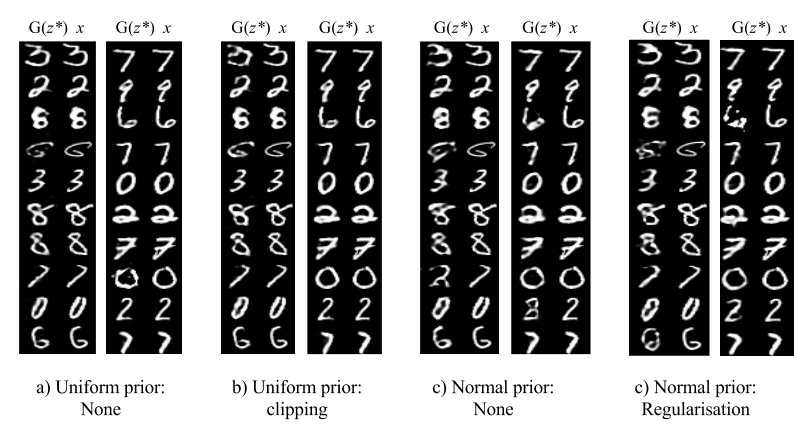}
    \caption{\textbf{Reconstructions for MNIST}: inverting a generator trained using a uniform prior (a-b) and a normal prior (c-d). The original image, $x$ is on the right, while the inverted image is on the left $G(z^*)$.}
    \label{rec_MNIST}
\end{figure}

Each MNIST digit is drawn in a unique style; a successful inversion of MNIST digits should preserve both the style and the identity of the digit. In Fig. \ref{rec_MNIST}, we show a random set of $20$ pairs of original images, $x$, and their reconstructions, $G(z^*)$. In general, the inversions preserve both style and identity well. Using visual inspection alone, it is not clear whether regularisation methods improve the inversion process or not. Table \ref{mnist_table} records the absolute, mean reconstruction error. Results suggest that the regularisation techniques that we employed did not improve the inversion. This is a positive result, as this suggests that inversion may be possible without regularisation, meaning that the inversion process can be independent of the noise process used. This also suggests that regions just outside of $P(Z)$ may also be able to produce meaningful examples from the data distribution.


\begin{table}
\centering
\begin{tabular}{ccccccc}\toprule
\multicolumn{3}{c}{\bf Uniform prior} &
\phantom{abc} &
\multicolumn{3}{c}{\bf Normal prior} \\ 
\cmidrule{1-3} \cmidrule{5-7} 
Without clipping & \phantom{abc} & With clipping & \phantom{abc} & Without regularisation & \phantom{abc} & With regularisation\\ 
\cmidrule{1-1} \cmidrule{3-3} \cmidrule{5-5} \cmidrule{7-7}
0.027 & \phantom{abc} & 0.032 & \phantom{abc} & 0.042 & \phantom{abc} & 0.044\\
\bottomrule
\end{tabular} 

\caption{\textbf{Reconstruction error for MNIST}: when inverting a generator trained using a Uniform or Normal prior, with and with out clipping or regularisation.}
\label{mnist_table}
\end{table}

\subsection{Omniglot}
\begin{figure}
    \centering
    \includegraphics[width=\textwidth]{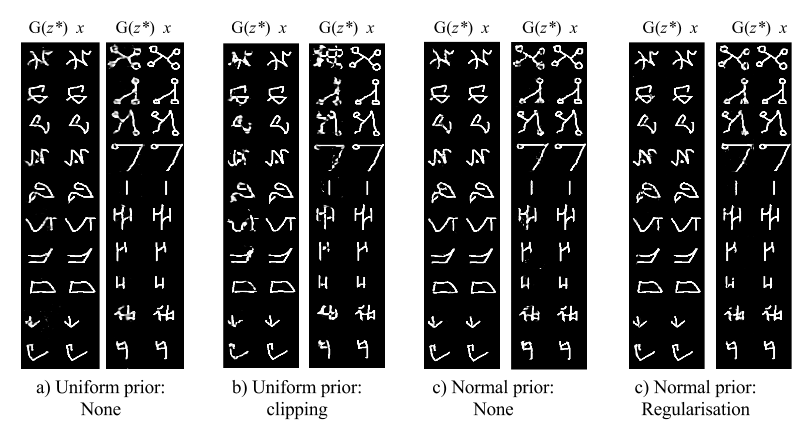}
    \caption{\textbf{Reconstructions for Omniglot}: inverting a generator trained using a uniform prior (a-b) and a normal prior (c-d). The generator is trained on a different set of alphabets to the testing samples.}\label{inv_omni}
\end{figure}

The Omniglot inversions are particularly challenging, as we are trying to find a set of $z^*$'s for a set of characters, $x$, from alphabets that were not in the training data. This is challenging the inversion process to invert samples from alphabets that it has not seen before, using information about alphabets that it has seen. The original and reconstructed samples are shown in Fig \ref{inv_omni}. In general, the reconstructions are sharp and able to capture fine details like small circles and edges. There is one severe fail case in Fig \ref{inv_omni} (b), where the top example has failed to invert the sample. A comparison of reconstruction error with and without regularisation is shown in Table \ref{omni_table}. These results suggest that regularisation does not improve inversion, and good inversion is possible without regularisation.


\begin{table}
\centering
\begin{tabular}{ccccccc}\toprule
\multicolumn{3}{c}{\bf Uniform prior} &
\phantom{abc} &
\multicolumn{3}{c}{\bf Normal prior} \\ 
\cmidrule{1-3} \cmidrule{5-7} 
Without clipping & \phantom{abc} & With clipping & \phantom{abc} & Without regularisation & \phantom{abc} & With regularisation\\ 
\cmidrule{1-1} \cmidrule{3-3} \cmidrule{5-5} \cmidrule{7-7}
0.023 & \phantom{abc} & 0.045 & \phantom{abc} & 0.020 & \phantom{abc} & 0.020 \\
\bottomrule
\end{tabular} 
\caption{\textbf{Reconstruction error for Omniglot}: when inverting a generator trained using a Uniform or Normal prior, with and with out clipping or regularisation.}
\label{omni_table}
\end{table}

\section{Conclusion}
The generator of a GAN learns the mapping $G: Z \rightarrow X$. It has been shown that $z$ values that are close in $Z$-space produce images that are visually similar in image space, $X$ \cite{radford2015unsupervised}. It has also been shown that images along projections in $Z$-space also have visual similarities \cite{radford2015unsupervised}. To exploit the structure of $Z$ for discriminative tasks, it is necessary to \textit{invert} this process, to obtain a latent encoding $z \in Z$ for an image $x \in X$. Inverting the generator also reveals interesting properties of the learned generative model.

We suggest a process for inverting the generator of any pre-trained GAN, obtaining a latent encoding for image samples, provided that the computational graph for the GAN is available. We presented candidate regularisation methods than can be used depending on the prior distribution over the latent space. However, we found that for the MNIST and Omniglot datasets that it is not necessary to use regularisation to perform the inversion, which means that this approach may be more generally applied. 

For GANs trained using batch normalisation, where only the gain and shift are learned but the mean and standard deviation are calculated on the go, it may not be possible to invert single image samples. If this is the case, it is necessary to invert batches of image samples. We show that it is indeed possible invert batches of image samples. Under reasonable assumptions, batch inversion is sensible because the gradient used to update a latent samples only depends on the reconstruction error of the latent sample that it is updating. Inverting batches may also make the inversion process more computationally efficient.



Our inversion results for the MNIST and Omniglot dataset provide interesting insight into the latent representation. For example, the MNIST dataset consists of handwritten digits, where each digit is written in a unique style. Our results also suggest that both the identity and the style of the digit is preserved in the inversion process we propose here, indicating that the latent space preserves both these properties. These results suggest that latent encodings may be useful for applications beyond digit classification. Results using the Omniglot dataset show that even handwritten characters from alphabets never seen during training of a GAN can be projected into the latent space, with good reconstructions. This may have implications for one-shot learning.




\section*{Acknowledgements}
We like to acknowledge the Engineering and Physical Sciences Research Council for funding through a Doctoral Training studentship.

\printbibliography

\end{document}